\documentclass[11pt,a4paper]{article}
\usepackage[hyperref]{acl2021}
\usepackage{times}
\usepackage{latexsym}

\usepackage{amsmath}
\usepackage{amssymb}
\usepackage{amsfonts}
\usepackage{enumitem}
\usepackage{graphicx}
\graphicspath{{Figure/}}

\usepackage{microtype}

\aclfinalcopy 
\def\aclpaperid{1943} 

\title{PLATO-2: Towards Building an Open-Domain Chatbot via\\ Curriculum Learning}

\author{Siqi Bao\thanks{~~Equal contribution.}~~~~ Huang He\footnotemark[1]~~~~ Fan Wang\footnotemark[1]~~~~ Hua Wu\footnotemark[1]~~~~ Haifeng Wang\\
	\bf{Wenquan Wu~~~~ Zhen Guo~~~~ Zhibin Liu~~~~ Xinchao Xu}\\
	Baidu Inc., China \\
	\texttt{\{baosiqi, hehuang, wang.fan, wu\_hua\}@baidu.com}}

\date{}

\begin{document}
\maketitle
\begin{abstract}
To build a high-quality open-domain chatbot, we introduce the effective training process of PLATO-2 via curriculum learning. There are two stages involved in the learning process. In the first stage, a coarse-grained generation model is trained to learn response generation under the simplified framework of one-to-one mapping. In the second stage, a fine-grained generative model augmented with latent variables and an evaluation model are further trained to generate diverse responses and to select the best response, respectively. PLATO-2 was trained on both Chinese and English data, whose effectiveness and superiority are verified through comprehensive evaluations, achieving new state-of-the-art results.
\end{abstract}

\section{Introduction}
Recently, task agnostic pre-training with large-scale transformer models has achieved great success in natural language processing \citep{devlin2019bert}, especially open-domain dialogue generation. For instance, based on the general language model GPT-2 \citep{radford2019language}, DialoGPT \citep{zhang2019dialogpt} is further trained for response generation using Reddit comments. To obtain a human-like open-domain chatbot, Meena \citep{adiwardana2020towards} scales up the network parameters to 2.6B and employs more social media conversations in the training process, leading to significant improvement on response quality. To mitigate undesirable toxic or bias traits of large corpora, Blender \citep{roller2020recipes} fine-tunes the pre-trained model with human annotated datasets and emphasizes desirable conversational skills of engagingness, knowledge, empathy and personality. 

In addition to the attempts from model scale and data selection, PLATO \citep{bao2019plato} aims to tackle the inherent one-to-many mapping problem to improve response quality. The one-to-many mapping refers to that one dialogue context might correspond to multiple appropriate responses. It is widely recognized that the capability of modeling one-to-many relationship is crucial for open-domain dialogue generation \citep{zhao2017learning, chen2019generating}. PLATO explicitly models this one-to-many relationship via discrete latent variables, aiming to boost the quality of dialogue generation. PLATO has a modest scale of 132M network parameters and trained with 8M samples, achieving relatively good performance among conversation models on a similar scale. However, scaling up PLATO directly encounters training instability and efficiency issues, which might result from the difficulty to capture the one-to-many semantic relationship from scratch. 

\begin{figure}
	\centering
	\includegraphics[width=0.95\columnwidth]{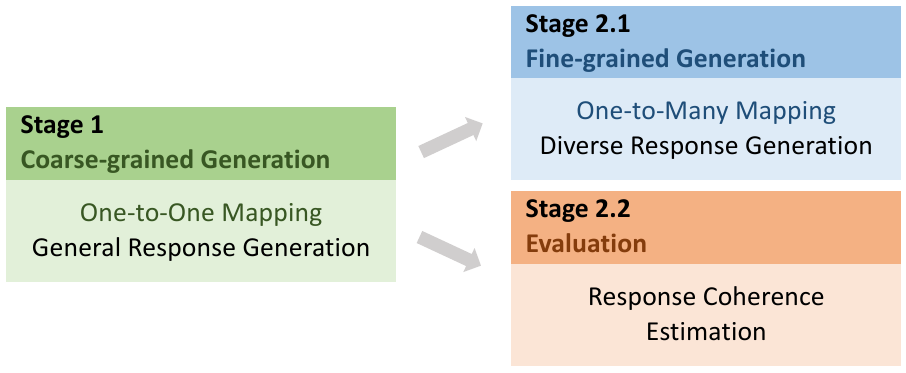}
	\caption{Curriculum learning process in PLATO-2.}
	\label{fig:overview}
\end{figure} 
\begin{figure*}
	\centering
	\includegraphics[width=0.99\textwidth]{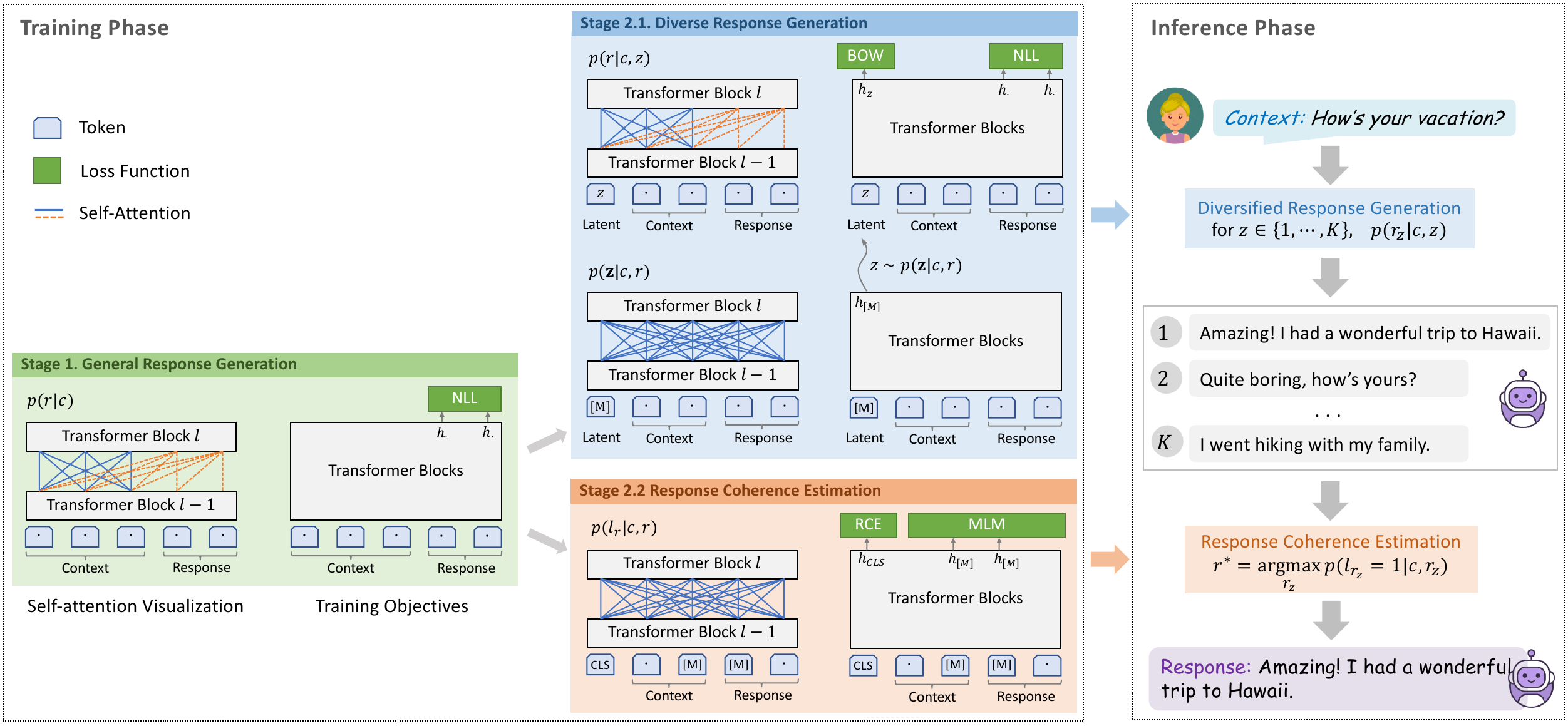}
	\caption{PLATO-2 illustration. Left: training phase via curriculum learning, model parameters in the second stage are warm started by those trained well in the first stage. Right: toy example to illustrate inference phase.}
	\label{fig:network}
\end{figure*} 
In this work, we try to scale up PLATO to PLATO-2 and introduce an effective training schema via curriculum learning \citep{bengio2009curriculum}. There are two stages involved in the whole learning process, as shown in Figure \ref{fig:overview}. In the first stage, under the simplified one-to-one mapping modeling, a \textbf{coarse-grained} generation model is trained for response generation under different conversation contexts. This model tends to capture typical patterns of diversified responses, sometimes resulting in general and dull responses during inference. Despite the problem of safe responses, this coarse-grained model is still highly effective in learning general concepts of response generation. 

The curriculum learning continues to the second stage, which contains the training of a \textbf{fine-grained} generation model and an evaluation model. The fine-grained generation model explicitly models the one-to-many mapping relationship via latent variables for diverse response generation. To select the most appropriate response, an evaluation model is trained to estimate the bi-directional coherence between the dialogue context and responses. Distinct with multi-task PLATO, the separate design of fine-grained generation and evaluation enables the model to concentrate more on its corresponding task, getting exempt from multi-task disturbance \citep{standley2019tasks}.

As compared with PLATO, PLATO-2 leverages curriculum learning to learn response generation gradually, from the general concept of one-to-one mapping to the complex concept of one-to-many mapping. With curriculum learning, we successfully scale the model up to billions of parameters, achieving new state-of-the-art results. Besides open-domain chitchat, the models learned in these two stages can also benefit task-oriented conversation and knowledge grounded dialogue respectively, whose effectiveness is verified thoroughly in DSTC9 \citep{gunasekara2020overview}.

To sum up, we trained PLATO-2 with different model sizes: 1.6B, 314M and 93M parameters. In addition to the English models, we also trained Chinese models with massive social media conversations. Comprehensive experiments on both English and Chinese datasets demonstrate that PLATO-2 outperforms Meena, Blender and other state-of-the-art models. We have released our English models and source codes at GitHub, hoping to facilitate the research in open-domain dialogue generation.\footnotemark[1]
\footnotetext[1]{\url{https://github.com/PaddlePaddle/Knover/tree/develop/projects/PLATO-2}}

\section{Methodology}
The backbone of PLATO-2 is consisted of transformer blocks with pre-normalization \citep{radford2019language}. Distinct with conventional Seq2Seq, there are no separate encoder and decoder networks in our infrastructure. PLATO-2 keeps the unified network for bi-directional context encoding and uni-directional response generation through flexible attention mechanism \citep{dong2019unified}. 

\subsection{Curriculum Learning}
In this work, we carry out effective training of PLATO-2 via curriculum learning. As shown in Figure \ref{fig:network}, there are two stages involved in the learning process: during stage 1, a coarse-grained baseline model is trained for general response generation under the simplified one-to-one mapping relationship; during stage 2, two models of fine-grained generation and evaluation are further trained for diverse response generation and response coherence estimation respectively. 

\subsubsection{General Response Generation}
It is well known that there exists a one-to-many relationship in open-domain conversations, where a piece of context may have multiple appropriate responses. Since conventional approaches try to fit the one-to-one mapping, they tend to generate generic and dull responses. Whereas, it is still an efficient way to capture the general characteristics of response generation. As such, we first train a coarse-grained baseline model to learn general response generation under the simplified relationship of one-to-one mapping. Given one training sample of context and response $(c, r)$, we need to minimize the following negative log-likelihood (NLL) loss:
\begin{equation}
\begin{split}
\mathcal{L}_{NLL}^{\text{Baseline}}&=-\mathbb{E}~\log p(r|c)\\
&=-\mathbb{E}~ \sum\nolimits_{t=1}^T~\log p(r_t|c,r_{<t})~,
\end{split}
\raisetag{2.6\baselineskip}
\end{equation}
where $T$ is the length of the target response $r$ and $r_{<t}$ denotes previously generated words. Since the response generation is a uni-directional decoding process, each token in the response only attends to those before it, shown as dashed orange lines in Figure \ref{fig:network}. As for the context, bi-directional attention is enabled for better natural language understanding, shown as blue lines in Figure \ref{fig:network}.

\subsubsection{Diverse Response Generation}
Based upon the coarse-grained baseline model, diverse response generation is warm started and further trained under the relationship of one-to-many mapping. Following the previous work PLATO, the discrete latent variable $z$ is introduced for the one-to-many relationship modeling. $z$ is one $K$-way categorical variable, with each value corresponding to a particular latent speech act in the response. The model will first estimate the latent act distribution of the training sample $p(\mathbf{z}|c,r)$ and then generate the response with the sampled latent variable $p(r|c,z)$. It is notable that these two tasks of response generation and latent act recognition are trained jointly within the shared network. The NLL loss of diverse response generation is defined as:
\begin{equation}
\begin{split}
\mathcal{L}_{NLL}^{\text{Generation}}&=-\mathbb{E}_{z\sim p(\mathbf{z}|c,r)} ~\log p(r|c,z)\\
&=-\mathbb{E}_{z\sim p(\mathbf{z}|c,r)} \sum\limits_{t=1}^T~\log p(r_t|c,z,r_{<t})~,
\end{split}
\raisetag{2.4\baselineskip}
\end{equation}
where $z$ is the latent act sampled from $p(\mathbf{z}|c,r)$. As sampling is not differentiable, we approximate it with Gumbel-Softmax \citep{jang2016categorical}. The posterior distribution over latent values is estimated through the task of latent act recognition: 
\begin{equation}
p(\mathbf{z}|c,r) =\text{softmax}(W_1 h_{[M]} + b_1) \in \mathbb{R}^K~,
\end{equation}
where $h_{[M]}\in \mathbb{R}^{D}$ is the final hidden state of the special mask token [M], $W_1\in \mathbb{R}^{K\times D}$ and $b_1\in \mathbb{R}^{K}$ denote the weight matrices of one fully-connected layer. 

To facilitate the training process of discrete latent variables, the bag-of-words (BOW) loss \citep{zhao2017learning} is also employed:
\begin{equation}
\begin{split}
\mathcal{L}_{BOW}^{\text{Generation}}&=-\mathbb{E}_{z\sim p(\mathbf{z}|c,r)}  \sum\nolimits_{t=1}^T ~ \log p(r_t|c,z)\\
&=-\mathbb{E}_{z\sim p(\mathbf{z}|c,r)} \sum\nolimits_{t=1}^T ~ \log \frac{e^{f_{r_t}}}{\sum\nolimits_{v\in V} e^{f_v}}~,
\end{split}
\raisetag{2.1\baselineskip}
\end{equation}
where $V$ refers to the whole vocabulary. The function $f$ tries to predict the words within the target response in a non-autoregressive way:
\begin{equation}
f=W_2 h_z+b_2 \in \mathbb{R}^{|V|}~,
\end{equation}
where $h_z$ is the final hidden state of the latent variable. $f_{r_t}$ denotes the estimated probability of word $r_t$. As compared with NLL loss, the BOW loss discards word orders and forces the latent variable to capture the global information of target response.

To sum up, the objective of the fine-grained generation model is to minimize the following integrated loss:
\begin{equation}
\mathcal{L}^{\text{Generation}}=\mathcal{L}_{NLL}^{\text{Generation}}+\mathcal{L}_{BOW}^{\text{Generation}}
\end{equation}

\subsubsection{Response Coherence Estimation}
By assigning distinct values to the latent variable, the fine-grained generation model is able to produce multiple high-quality and diverse responses. To select the most appropriate response from these candidates, one straightforward way is to rank them according to $p(z|c)p(r|c,z)$. However, it is widely recognized that the prior distribution $p(\mathbf{z}|c)$ is difficult to estimate and the uniform distribution is not an effective approximation. To this end, we adopt an alternative approach to train an evaluation model in the second stage, estimating the coherence between each response and the given dialogue context. The loss of response coherence estimation (RCE) is defined as follows:
\begin{equation}
\begin{split}
\mathcal{L}_{RCE}^{\text{Evaluation}}=&-\log p(l_r=1|c,r)\\
&-\log p(l_{r^-}=0|c,r^-)
\end{split}
\end{equation}
The positive training samples come from the dialogue context and corresponding target response $(c,r)$, with coherence label $l_r=1$. And the negative samples are created by randomly selecting responses from the corpus $(c,r^-)$, with coherence label $l_{r^-}=0$.

In addition to our coherence evaluation function $p(l_r|c,r)$, there are two other functions widely used for response selection. One is the length-average log-likelihood \citep{adiwardana2020towards}, which considers the forward response generation probability $p(r|c)$. The other one is the maximum mutual information \citep{zhang2019dialogpt}, which considers the backward context recovery probability $p(c|r)$. However, the forward score favors safe and generic responses due to the property of maximum likelihood, while the backward score tends to select the response with a high overlap with the context, resulting in repetitive conversations. By contrast, the discriminative function $p(l_r|c,r)$ considers the bi-directional information flow between the dialogue context and response. Our coherence evaluation is able to ameliorate the aforementioned problems, whose effectiveness is verified in the experiments.

To maintain the capacity of distributed representation, the task of masked language model (MLM) \citep{devlin2019bert} is also included in the evaluation network. Within this task, 15\% of the input tokens will be masked at random and the network needs to recover the masked ones. The MLM loss is defined as:
\begin{equation}
\mathcal{L}_{MLM}^{\text{Evaluation}}=-\mathbb{E} \sum\nolimits_{m\in M} \log p(x_m|x_{\backslash M}),
\end{equation}
where $x$ refers to the input tokens of context and response. $\{x_m\}_{m\in M}$ stands for masked tokens and $x_{\backslash M}$ denotes the rest unmasked ones.

To sum up, the objective of the evaluation model is to minimize the following integrated loss:
\begin{equation}
\mathcal{L}^{\text{Evaluation}}=\mathcal{L}_{RCE}^{\text{Evaluation}}+\mathcal{L}_{MLM}^{\text{Evaluation}}
\end{equation}

\subsection{Inference} 
For open-domain chitchat, the inference is carried out with the second stage's models as follows.
\begin{enumerate}[label=\arabic*),leftmargin=*,noitemsep,topsep=0pt]
	\item Diverse response generation. Conditioned on each latent value $z\in \{1,\cdots,K\}$, its corresponding candidate response $r_z$ is produced by the fine-grained generation model $p(r_z|c, z)$. 

    \item Response coherence estimation. The evaluation model will preform ranking and select the one with highest coherence value as the final response $r^* = \text{argmax}_{r_z}~ p(l_{r_z}=1|c,r_z)$.
\end{enumerate}

\section{Experiments}
\subsection{Training Data}
PLATO-2 has English and Chinese models, with training data extracted from open-domain social media conversations. The English training data is extracted from Reddit comments, which are collected by a third party and made publicly available on pushshift.io \citep{baumgartner2020pushshift}. To improve the generation quality, we carry out elaborate data cleaning, as discussed in the Appendix. After filtering, the data is split into training and validation sets in chronological order. The training set contains 684M (context, response) samples, ranging from December 2005 to July 2019. For the validation set, 0.2M samples are selected from the rest data after July 2019. The English vocabulary contains 8K BPE tokens \citep{sennrich2016neural}, constructed with the SentencePiece library.

The Chinese training data is collected from public domain social medias. After filtering, there are 1.2B (context, response) samples in the training set, 0.1M samples in the validation set, and 0.1M samples in the test set. As for the Chinese vocabulary, it contains 30K BPE tokens.

\subsection{Training Details}
PLATO-2 has three model sizes: a standard version of 1.6B parameters, a small version of 314M parameters, and a tiny version of 93M parameters. Detailed network and training configurations are summarized in the Appendix. The main hyper-parameters used in the training process are listed as follows. The maximum sequence lengths of context and response are all set to 128. $K$ is set to 20 for the discrete latent variable \citep{bao2019plato, chen2019generating}. We use Adam \citep{kingma2015adam} as the optimizer, with a learning rate scheduler including a linear warmup and an invsqrt decay \citep{vaswani2017attention}. To train the large-scale model with a relatively large batch size, we employ gradient checkpointing \citep{chen2016training} to trade computation for memory. The training was carried out on 64 Nvidia Tesla V100 32G GPU cards. It takes about 3 weeks for 1.6B parameter model to accomplish curriculum learning process.

\subsection{Evaluation Settings}
\subsubsection{Compared Methods}
The following methods have been compared in the experiments.
\begin{itemize}[leftmargin=*,noitemsep,topsep=0pt]
    \item PLATO \citep{bao2019plato} is trained on the basis of BERT\textsubscript{BASE} using 8.3M Twitter and Reddit conversations \citep{cho2014learning, zhou2018commonsense, galley2019grounded}. There are 132M network parameters in this model. 

	\item DialoGPT \citep{zhang2019dialogpt} is trained on the basis of GPT-2 \citep{radford2019language} using Reddit comments. There are three model sizes: 117M, 345M and 762M. Since the 345M parameter model obtains the best performance in their evaluations, we compare with this version.
	
	\item Blender \citep{roller2020recipes} is first trained using Reddit comments and then fine-tuned with human annotated conversations -- BST \citep{smith2020can}, to help emphasize desirable conversational skills of engagingness, knowledge, empathy and personality. Blender has three model sizes: 90M, 2.7B and 9.4B. Since the 2.7B parameter model obtains the best performance in their evaluations, we compare with this version.
	
	\item Meena \citep{adiwardana2020towards} is an open-domain chatbot trained with social media conversations. There are 2.6B network parameters in Meena. Since Meena has not released the model or provided a service interface, it is difficult to perform comprehensive comparison. In the experiments, we include the provided samples in their paper for static evaluation.
	
	\item Microsoft XiaoIce \citep{zhou2020design} is a popular social chatbot in Chinese. The official Weibo platform is used in the evaluation.
\end{itemize}

For the sake of comprehensive and fair comparisons, different versions of PLATO-2 are included in the experiments.
\begin{itemize}[leftmargin=*,noitemsep,topsep=0pt]
	\item PLATO-2 1.6B parameter model is the standard version in English, which is first trained using Reddit comments and then fine-tuned with BST conversations. To measure the effectiveness of PLATO-2, this model will be compared to the state-of-the-art open-domain chatbot Blender.
	
	\item PLATO-2 314M parameter model is a small version in English, which is trained with Reddit comments. This model will be compared to DialoGPT, as they have similar model scales.
	
	\item PLATO-2 93M parameter model is a tiny version in English, which is trained with Reddit comments. As it is difficult to scale up PLATO, we use this version to compare with PLATO.
	
	\item PLATO-2 336M parameter Chinese model\footnotemark[2] will be compared to XiaoIce in the experiments.
\end{itemize}
\footnotetext[2]{This model has 24 transformer blocks and 16 attention heads, with the embedding dimension of 1024. As the Chinese vocabulary contains 30K BPE tokens, this model has 22.5M more parameters than the English small model.}

\begin{table*}[ht]
	\centering
	\includegraphics[width=\textwidth]{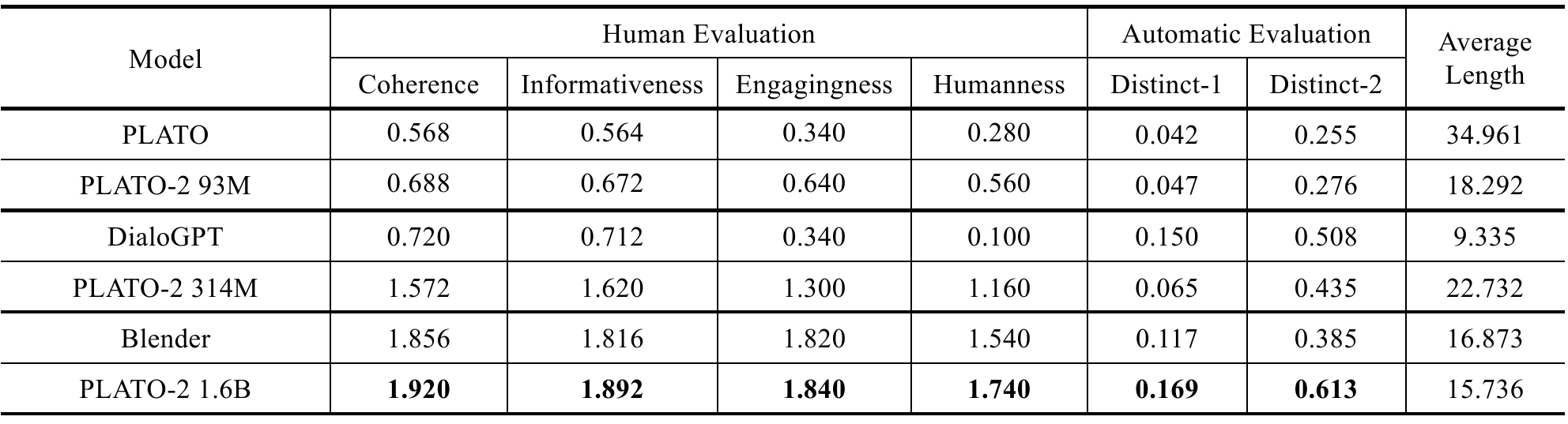}
	\caption{Self-chat evaluation results, with best value written in bold.}
	\label{tab:self-chat}
\end{table*} 
\begin{table*}[ht]
	\centering
	\includegraphics[width=\textwidth]{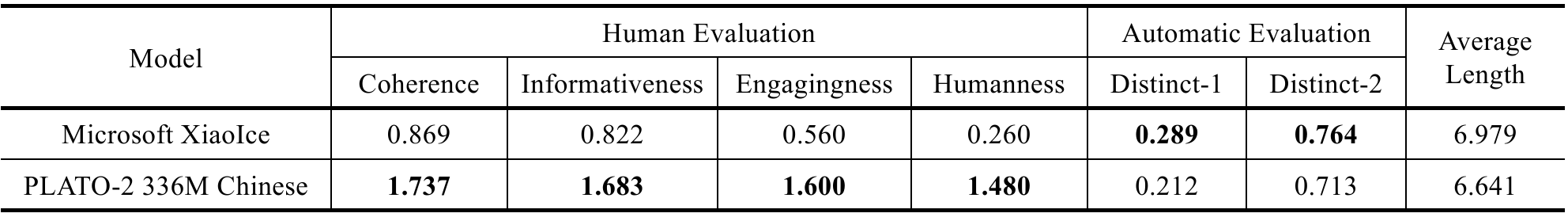}
	\caption{Chinese interactive evaluation results, with best value written in bold.}
	\label{tab:interactive}
\end{table*} 
\begin{figure*}[ht]
	\centering
	\includegraphics[width=\textwidth]{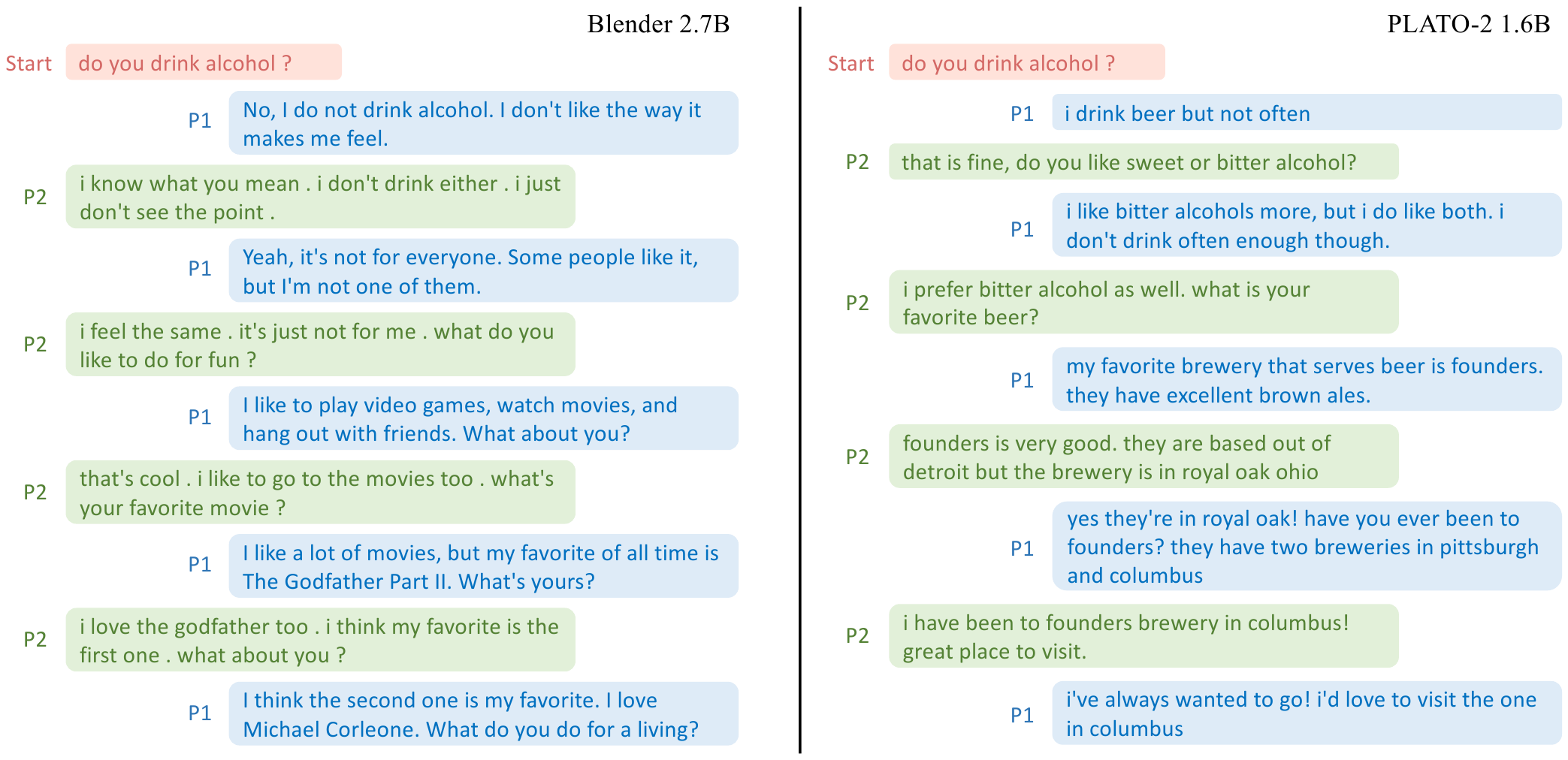}
	\caption{Self-chat examples by Blender and PLATO-2.}
	\label{fig:case}
\end{figure*} 

\subsubsection{Evaluation Metrics}
We carry out both automatic and human evaluations in the experiments. In automatic evaluation, to assess the model's capacity on lexical diversity, we use the corpus-level metric of distinct-1/2 \citep{li2016diversity}, which is defined as the number of distinct uni- or bi-grams divided by the total number of generated words.

In human evaluation, we employ four utterance-level and dialogue-level metrics, including coherence, informativeness, engagingness and humanness. Three crowd-sourcing workers are asked to score the response/dialogue quality on a scale of [0, 1, 2], with the final score determined through majority voting. The higher score, the better. These criteria are discussed as follows, with scoring details provided in the Appendix.
\begin{itemize}[leftmargin=*,noitemsep,topsep=0pt]
	\item Coherence is an utterance-level metric, measuring whether the response is relevant and consistent with the context.
	
	\item Informativeness is also an utterance-level metric, evaluating whether the response is informative or not given the context.
	
	\item Engagingness is a dialogue-level metric, assessing whether the annotator would like to talk with the speaker for a long conversation.
	
	\item Humanness is also a dialogue-level metric, judging whether the speaker is a human being or not.
\end{itemize}

\subsection{Experimental Results}
In the experiments, we include both static and interactive evaluations. 

\subsubsection{Self-Chat Evaluation}
Self-chats have been widely used in the evaluation of dialogue systems \citep{li2016deep,bao2019know,roller2020recipes}, where a model plays the role of both partners in the conversation. As compared with human-bot conversations, self-chat logs can be collected efficiently at a cheaper price. As reported in \citet{li2019acute}, self-chat evaluations exhibit high agreement with the human-bot chat evaluations. In the experiments, we ask the bot to perform self-chats and then invite crowd-sourcing workers to evaluate the dialogue quality.

The way to start the interactive conversation needs special attention. As pointed out by \citet{roller2020recipes}, if starting with `Hi!', partners tend to greet with each other and only cover some shallow topics in the short conversation. Therefore, to expose the model's weaknesses and explore the model's limits, we choose to start the interactive conversation with pre-selected topics. We use the classical 200 questions as the start topic \citep{vinyals2015neural} and ask the bot to performance self-chats given the context. There are 10 utterances in each dialogue, including the input start utterance. We carry out automatic evaluation on the 200 self-chat logs and randomly select 50 conversations for human evaluation. 

The compared models are divided into three groups. The first group includes PLATO 132M model and PLATO-2 93M model. Both of them have similar model scales. The second group includes DialoGPT 345M model and PLATO-2 310M model. Both of them are trained using Reddit comments and have similar model scales. The third group includes Blender 2.7B model and PLATO-2 1.6B model. Both of them are first trained using Reddit comments and further fine-tuned with BST conversations. In human evaluation, two self-chat logs, which are from the same group and have the same start topic, will be displayed to three annotators. One example is given in Figure \ref{fig:case}. As suggested in ACUTE-Eval \citep{li2019acute}, we ask crowd-sourcing workers to pay attention to only one speaker within a dialogue. In the evaluation, they need to give scores on coherence and informativeness for each P1's utterance, and assess P1's overall quality on engagingness and humanness. 

The self-chat evaluation results are summarized in Table \ref{tab:self-chat}. These results indicate that PLATO-2 1.6B model obtains the best performance across human and automatic evaluations. In the first group, PLATO-2 achieves better performance than PLATO on a similar model scale, which might mainly result from the stable curriculum learning and large-scale conversation data. In the second group, DialoGPT tends to generate repetitive conversations due to the backward scoring function, resulting in poor performance in interactive evaluation. In the third group, PLATO-2 outperforms the state-of-the-art open-domain chatbot Blender. The gap of Blender and PLATO-2 on the corpus-level metric distinct-1/2 suggests that PLATO-2 has a better capacity on lexical diversity. In addition, the difference among these three groups suggests that enlarging model scales and exploiting human annotated conversations help improve the dialogue quality.

\subsubsection{Human-Bot Chat Evaluation}
In the Chinese evaluation, it is difficult to carry out self-chats for Microsoft XiaoIce, as there is no public available API. Therefore, we collect human-bot conversations through their official Weibo platform. The interactive conversation also starts with a pre-selected topic and continues for 7-14 rounds. 50 diverse topics are extracted from the high-frequency topics of a commercial chatbot, including travel, movie, hobby and so on. The collected human-bot conversations are distributed to crowd-sourcing workers for evaluation. The human and automatic evaluation results are summarized in Table \ref{tab:interactive}. XiaoIce obtains higher distinct values, which may use a retrieval-based strategy in response generation. The human evaluations indicate that our PLATO-2 model achieves significant improvements over XiaoIce across all the human evaluation metrics. 

\subsubsection{Static Evaluation}
Besides the interactive evaluation, we also employ static evaluation to analyze the model's performance. In static evaluation, each model will produce a response towards the given multi-turn context. Those powerful models are involved in the evaluation: Meena, Blender, DialoGPT and PLATO-2 1.6B. To compare with Meena, we include their provided 60 static samples in the Appendix of the paper and generate corresponding responses with other models. We also include 60 test samples about daily life from Daily Dialog \citep{li2017dailydialog} and 60 test samples about in-depth discussion from Reddit. Given that the measurement of humanness usually needs multi-turn interaction, this metric is excluded from static evaluation. The evaluation results are summarized in Table \ref{tab:static_1}. It can be observed that PLATO-2 is able to produce coherent, informative and engaging responses across different chat scenarios. The average Fleiss’s kappa \citep{fleiss1971measuring} of human evaluation is 0.466, indicating annotators have reached moderate agreement.
\begin{table}[ht]
	\centering
	\includegraphics[width=0.48\textwidth]{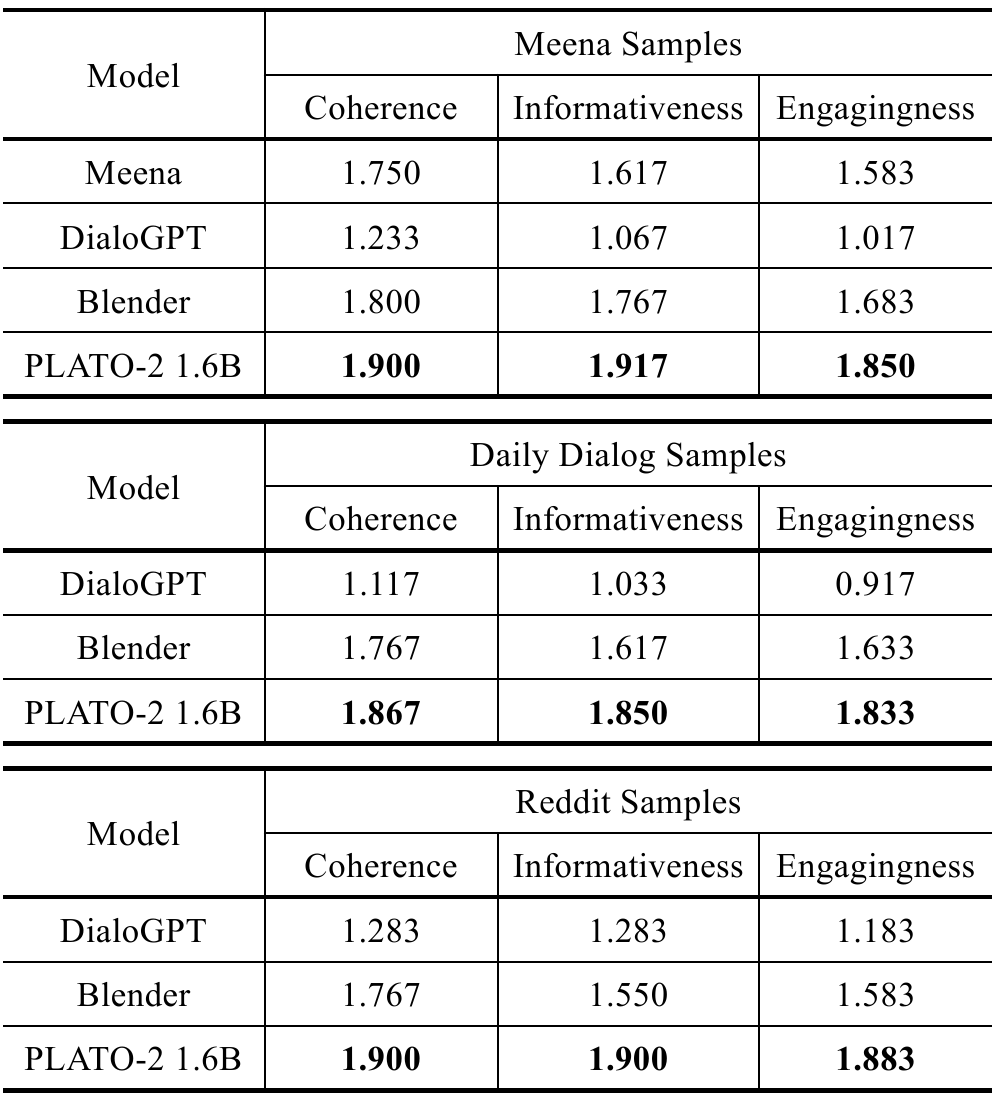}
	\caption{Static evaluation results, with the best scores written in bold.}
	\label{tab:static_1}
\end{table}

\subsection{Discussions}
\subsubsection{Case Analysis}
To further analyze the models' features, two self-chat examples of Blender and PLATO-2 are provided in Figure \ref{fig:case}. Although both models are able to produce high-quality engaging conversations, they exhibit distinct discourse styles. Blender tends to switch topics quickly in the short conversation, including alcohol, hobbies, movies and work. The emergence of this style might be related with BST fine-tuning data. For instance, persona chat in BST is about the exchange of personal information between two partners, where topics need to switch quickly to know more about each other. Due to the task settings of data collection, some human annotated conversations might be a little unnatural. Nevertheless, fine-tuning with BST conversations is essential to mitigate undesirable toxic traits of large corpora and emphasize desirable skills of human conversations. 

Distinct with Blender, PLATO-2 can stick to the start topic and conduct in-depth discussions. The reasons might be two-fold. First, our model is able to generate diverse and informative responses with the accurate modeling of one-to-many relationship. Second, the evaluation model helps select the coherent response and stick to current topic. We asked crowd-sourcing workers to annotate which model's in-depth discussion is better w.r.t. the start topic. The comparison result is shown in Table \ref{tab:thorough}, which also verifies our above analysis on discourse styles. 

\begin{table}
	\centering
	\includegraphics[width=0.48\textwidth]{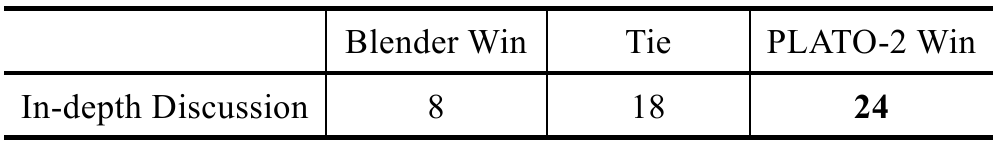}
	\caption{In-depth discussion w.r.t. the start topic.}
	\label{tab:thorough}
\end{table} 
\subsubsection{Why PLATO-2 Performs Better?}
Why PLATO-2 achieves better performance as compared with Meena, Blender and other state-of-the-art models? As analyzed above, major reasons might come from two aspects: fine-grained generation and evaluation. First, PLATO-2 employs discrete latent variable for the one-to-many relationship modeling, which is able to generate high-quality and diverse responses. Second, the evaluation model in PLATO-2 is effective at selecting the most appropriate response from the candidates.

\begin{table}
	\centering
	\includegraphics[width=0.48\textwidth]{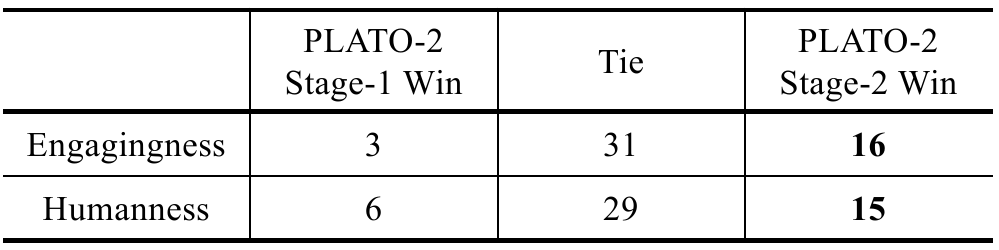}
	\caption{Comparison of the models in PLATO-2.}
	\label{tab:ablation}
\end{table} 
In fact, these two aspects are associated with the curriculum learning in the second stage, modeling the one-to-many relationship for open-domain conversations. By contrast, Meena and Blender are learned under the one-to-one mapping relationship, similar to the first stage in PLATO-2. To dissect the effects of these two stage models, we further ask crowd-sourcing workers to evaluate the models' self-chat logs on the dialogue-level metrics. The comparison results are summarized in Table \ref{tab:ablation}. These results verify the effectiveness of curriculum learning in PLATO-2. 

\subsubsection{Further Exploration of PLATO-2}
In addition to open-domain chitchat, there are two other kinds of dialogues in conversational AI \citep{gao2018neural}: knowledge grounded dialogue, and task-oriented conversation. Similar to open-domain conversation, the one-to-many mapping relationship also exists in knowledge grounded dialogue \citep{kim2019sequential}: given a dialogue context, multiple pieces of knowledge might be applicable for the response generation. Therefore, the one-to-many mapping models of the second stage can also be adapted for knowledge grounded dialogue. By expanding the network input with the knowledge segment, the background knowledge is encoded and grounded for response generation. Distinct from the open-domain conversation and knowledge grounded dialogue, task-oriented conversations usually need to accomplish a specific goal. Accordingly, the conversation flow would become less diverse and concentrated on task completion. Therefore, the one-to-one mapping generation model of the first stage can be used for the end-to-end task-oriented conversation.

For the exploration of PLATO-2 two-stage framework, we participated in several tasks of DSTC9 \citep{gunasekara2020overview}, including interactive evaluation of open-domain conversation (Track3-task2), static evaluation of knowledge grounded dialogue (Track3-task1), and end-to-end task-oriented conversation (Track2-task1). PLATO-2 has achieved the first place in all three tasks \citep{bao2021unified}. To sum up, the benefits brought by the two-stage curriculum learning in PLATO-2 are two-fold. Firstly, given the difficulties to scale up PLATO, the two-stage curriculum learning is an essential ingredient for the successful training of 1.6B parameter PLATO-2. Secondly, the two-stage PLATO-2 adapts well to multiple conversational tasks, indicating its potentials as a unified pre-training framework for conversational AI.

\section{Related Work}
Related works include large-scale language models and open-domain dialogue generation.

\noindent \textbf{Large-scale Language Models.} Pre-trained large-scale language models have brought many breakthroughs on various NLP tasks. GPT \citep{radford2018improving} and BERT \citep{devlin2019bert} are representative uni-directional and bi-directional language models, trained on general text corpora. By introducing pre-normalization and modifying weight initialization, GPT-2 \citep{radford2019language} successfully extends the model scale from 117M to 1.5B parameters. To cope with memory constraints, Megatron-LM \citep{shoeybi2019megatron} exploits model parallelism to train an 8.3B parameter model on 512 GPUs. GPT-3 \citep{brown2020language} further trains an 175B parameter autoregressive language model, demonstrating strong performance on many NLP tasks. The development of large-scale language models is also beneficial to the task of dialogue generation. 

\noindent \textbf{Open-domain Dialogue Generation.} On the basis of GPT-2, DialoGPT \citep{zhang2019dialogpt} is trained for response generation using Reddit comments. To obtain a human-like open-domain chatbot, Meena \citep{adiwardana2020towards} scales up the network parameters to 2.6B and utilizes more social media conversations in the training process. To emphasize desirable conversational skills of engagingness, knowledge, empathy and personality, Blender \citep{roller2020recipes} further fine-tunes the pre-trained model with human annotated conversations. In addition to the attempts on model scale and data selection, PLATO introduces discrete latent variable to tackle the inherent one-to-many mapping problem to improve response quality. In this work, we explore the effective training of PLATO-2 via curriculum learning.

\section{Conclusion}
In this work, we discuss the effective training of open-domain chatbot PLATO-2 via curriculum learning, where two stages are involved. In the first stage, one coarse-grained model is trained for general response generation. In the second stage, two models of fine-grained generation and evaluation are trained for diverse response generation and response coherence estimation. Experimental results demonstrate that PLATO-2 achieves substantial improvements over the state-of-the-art methods in both Chinese and English evaluations.  

\section*{Acknowledgments}
We would like to thank Jingzhou He, and Tingting Li for the help on resource coordination; Daxiang Dong, and Pingshuo Ma for the support on PaddlePaddle; Yu Sun, Yukun Li, and Han Zhang for the assistance with infrastructure and implementation. This work was supported by the Natural Key Research and Development Project of China (No. 2018AAA0101900).
	
\bibliographystyle{acl_natbib}
\bibliography{acl2021}

\appendix
\section{Data Cleaning Process}
PLATO-2 has English and Chinese models, with training data extracted from open-domain social media conversations. As the comments are formatted in message trees, any conversation path from the root to a tree node can be treated as one training sample, with the node as response and its former turns as context. To improve the generation quality, we carry out elaborate data cleaning. A message node and its sub-trees will be removed if any of the following conditions is met.
\begin{enumerate}[label=\arabic*),noitemsep,topsep=0pt]
	\item The number of BPE tokens is more than 128 or less than 2.
	\item Any word has more than 30 characters or the message has more than 1024 characters.
	\item The percentage of alphabetic characters is less than 70\%.
	\item The message contains URL.
	\item The message contains special strings, such as r/, u/, \&amp.
	\item The message has a high overlap with the parent's text.
	\item The message is repeated more than 100 times.
	\item The message contains offensive words.
	\item The subreddit is quarantined.
	\item The author is a known bot.
\end{enumerate}

After data cleaning, the English training data contains 684M (context, response) samples and the Chinese training data contains 1.2B (context, response) samples. Each English/Chinese sample has 2.78/2.82 utterances and each utterance has 26.29/22.20 tokens on average.

\section{Training Configurations}
PLATO-2 has three model sizes: a standard version of 1.6B parameters, a small version of 314M parameters, and a tiny version of 93M parameters. The 1.6B parameter model has 32 transformer blocks and 32 attention heads, with the embedding dimension of 2048. The 314M parameter model has 24 transformer blocks and 16 attention heads, with the embedding dimension of 1024. The 93M parameter model has 12 transformer blocks and 12 attention heads, with the embedding dimension of 768. The training configurations of PLATO-2 1.6B are provided in Table \ref{tab:training}. The training was carried out on 64 Nvidia Tesla V100 32G GPU cards. It takes about 3 weeks for the 1.6B parameter model to accomplish the curriculum learning process.

\begin{table*}
	\centering
	\includegraphics[width=\textwidth]{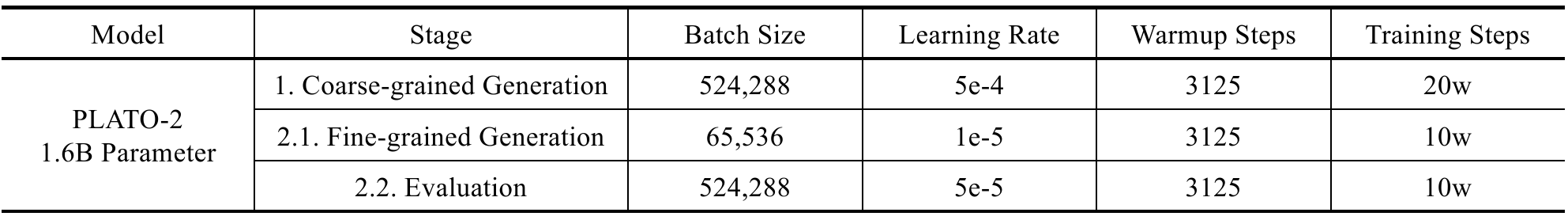}
	\caption{Training configurations of PLATO-2.}
	\label{tab:training}
\end{table*} 
\begin{figure*}
	\centering
	\includegraphics[width=\textwidth]{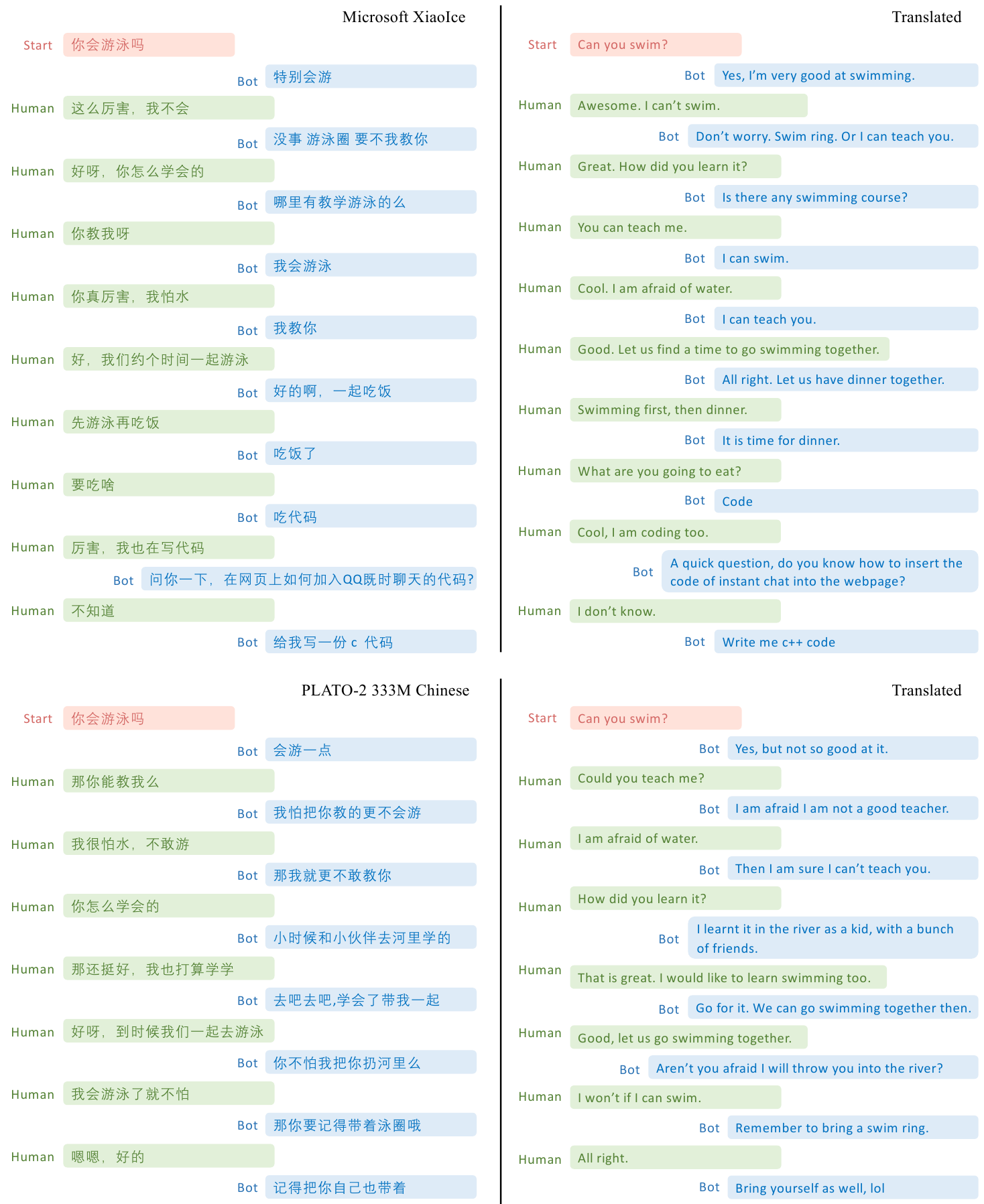}
	\caption{Human-bot chat examples by Microsoft XiaoIce and PLATO-2.}
	\label{fig:case_zh}
\end{figure*} 

\section{Chinese Case Analysis}
We also provide two human-bot chat examples of XiaoIce and PLATO-2 in Figure \ref{fig:case_zh}, with original interactive logs shown on the left and translated logs on the right. It can be observed that some responses produced by XiaoIce are not coherent with the contexts and there are some abrupt changes of topics. By contrast, the interaction with PLATO-2 is more coherent and engaging.

\section{Scoring Criteria in Human Evaluation}
The criteria used in human evaluation are provided in Table \ref{tab:criteria}.
\begin{table}[ht]
	\centering
	\includegraphics[width=0.48\textwidth]{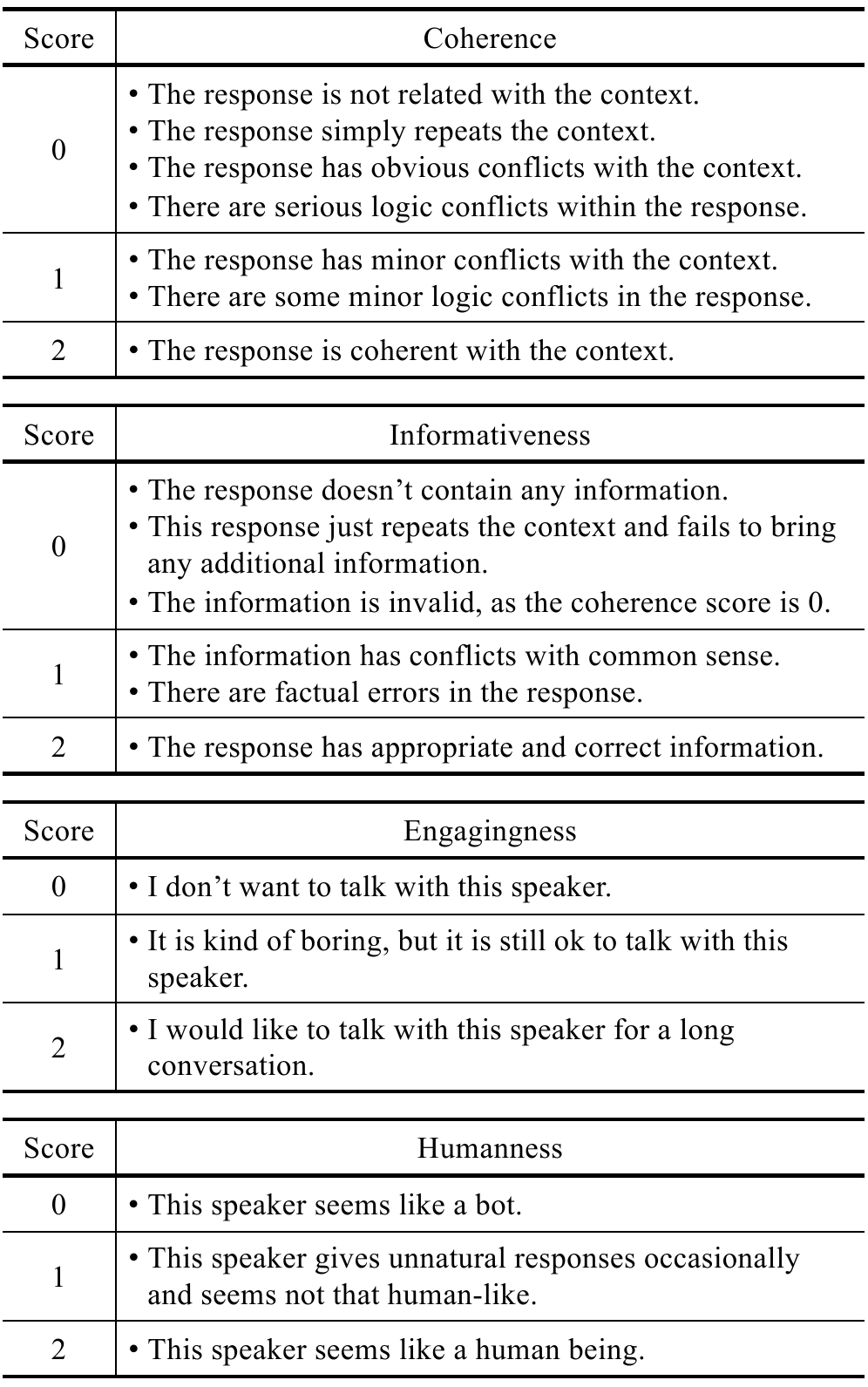}
	\caption{Score details of four metrics in human evaluation.}
	\label{tab:criteria}
\end{table}
\begin{table}
	\centering
	\includegraphics[width=0.48\textwidth]{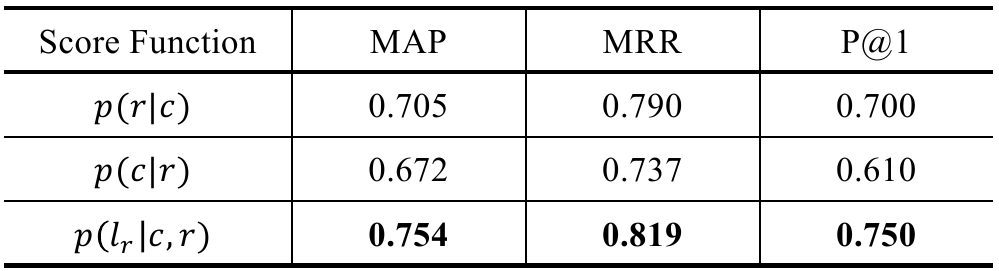}
	\caption{Comparison of different score functions in response selection, with best value written in bold.}
	\label{tab:selection}
\end{table} 

\section{Response Selection Comparison}
We carry out more experiments to compare the performance of distinct scoring functions in response selection. Firstly, one Chinese response selection dataset is constructed: 100 dialogue contexts are selected from the test set and 10 candidate responses are retrieved for each context with a commercial chatbot. Secondly, we ask crowd-sourcing workers to annotate the label whether the candidate response is coherent with the context. Thirdly, we train three 336M parameter models as the scoring function, including the forward response generation probability $p(r|c)$, the backward context recover probability $p(c|r)$ and the bi-directional coherence probability $p(l_r|c,r)$. Their results on the annotated response selection dataset are summarized in Table \ref{tab:selection}. The metrics of mean average precision (MAP), mean reciprocal rank (MRR) and precision at position 1 (P@1) are employed. These results indicate that PLATO-2's evaluation model is better at selecting appropriate responses.

\end{document}